\documentclass[conference]{IEEEtran}
\usepackage[T1]{fontenc}
\usepackage{newtxtext,newtxmath}
\usepackage{booktabs}
\usepackage{graphicx}
\usepackage{cite}
\usepackage{amsmath}
\usepackage{url}
\usepackage[hidelinks]{hyperref}
\renewenvironment{abstract}{\small\bfseries\noindent\textit{Abstract:}\ }{\par}
\renewenvironment{IEEEkeywords}{\small\bfseries\noindent\textit{Index Terms:}\ }{\par\vspace{4pt}}

\begin{document}

\title{A Multi-Word Tokenizer for 204 Languages That Needs Fewer Tokens Than GPT-4o's, Including on English}

\author{\IEEEauthorblockN{Johin Johny Arimbur}
\IEEEauthorblockA{Havon AI\\
Mumbai, India\\
johinjohny144@gmail.com}
\and
\IEEEauthorblockN{Ankur Ashit Ghosh}
\IEEEauthorblockA{Havon AI\\
Mumbai, India\\
ankurghosh2910@gmail.com}
\and
\IEEEauthorblockN{Shervin Sam}
\IEEEauthorblockA{Havon AI\\
Mumbai, India\\
samshervin37@gmail.com}}

\maketitle

\begin{abstract}
Languages of India need 1.6 to 14 times as many tokens as English on GPT-4o's tokenizer for the same content, which makes them more expensive and slower to serve. We train tokenizers from scratch on human-written text in about 200 languages. Havon v3 is a byte-level BPE tokenizer with 262,144 tokens and a pre-tokenizer that keeps combining marks with their letters and chunks scripts written without spaces. On FLORES-200 devtest it needs fewer tokens than GPT-4o's o200k\_base on all 25 languages of India (median 0.74 of GPT-4o's count) and on 172 of 204 languages, but 7\% more on English. Havon v4 adds a second, multi-word training stage (SuperBPE) to v3. At exactly GPT-4o's vocabulary size of 200,019 tokens, v4 needs 6.1\% fewer tokens than GPT-4o on English and fewer on 161 of 204 languages and all 25 languages of India (median 0.74); at the same size, v3 needs 9.4\% more on English and wins on 133. With 262,144 tokens, v4 needs 9.6\% fewer tokens on English and fewer on 182 of 204 languages, and with 524,288 tokens it needs fewer on 203 of 204. The gains are largest for English, Hindi and the Romance languages; scripts without spaces, such as Khmer, Burmese and Tibetan, gain little and lose subword merges. Havon v5 replaces v4's Python-only code sample with permissively licensed code in 15 languages: on code it needs 1.03 to 1.22 times GPT-4o's tokens, against 1.11 to 1.29 for v4, while keeping all 25 languages of India (median 0.74, all significant under a sentence-level bootstrap) and English (0.94). These are compression results only: we have not yet evaluated a model trained with v4 or v5.
\end{abstract}


\section{Introduction}
Language model APIs bill per token and context windows are measured in tokens, so a tokenizer that splits one language into more pieces makes that language more expensive, slower and shorter in context. In a companion measurement study~\cite{johny2026tax} we found that, on GPT-4o's tokenizer, the 25 languages of India in FLORES-200 need between 1.57 times (Hindi) and 13.7 times (Santali in Ol Chiki) the tokens of the same English content, and that Sarvam-30B/105B, the most even public tokenizer, still needs 1.21 to 2.50 times.

This paper asks whether one tokenizer can remove most of this tax without giving up English or the world's other languages. GPT-4o's o200k\_base is a demanding reference: it is the tokenizer of OpenAI's current models and is strong on English and on the major European and East Asian languages. We make five contributions:
\begin{enumerate}
\item Havon v3, a 262,144-token byte-level BPE tokenizer trained from scratch on human-written text in about 200 languages, with a pre-tokenizer that keeps Indic vowel signs with their letters and bounds chunks in scripts written without spaces.
\item Havon v4, which continues v3's training with multi-word merges (SuperBPE~\cite{liu2025superbpe}), at 200,019, 262,144 and 524,288 tokens.
\item An equal-vocabulary comparison: at exactly GPT-4o's 200,019 tokens, v4 needs fewer tokens than GPT-4o on English and on 161 of 204 FLORES-200 languages.
\item Controls that separate the multi-word stage from vocabulary size and data: v3 cut to 200,019 tokens, a 40,000-token test on identical text, and a data-budget ablation.
\item Havon v5, which extends the recipe to code, and a comparison with 12 public tokenizers, including Llama~3 and~4, Gemma~2 and~3, BLOOM and Mistral NeMo (Fig.~\ref{fig:tokenizers}).
\end{enumerate}

\section{Related Work}
Byte-pair encoding~\cite{sennrich2016bpe} and its byte-level form~\cite{radford2019gpt2} underlie most current tokenizers. Petrov et al.~\cite{petrov2023unfairness} and Ahia et al.~\cite{ahia2023allcost} showed that tokenizers impose very different costs across languages, and Rust et al.~\cite{rust2021tokenizer} that tokenizer quality affects multilingual model performance. Tamang and Bora~\cite{tamang2024indian} compared 12 tokenizers on the 22 official languages of India. MorphTok~\cite{brahma2025morphtok} proposed morphology-aware pre-tokenization for Hindi and Marathi. SuperBPE~\cite{liu2025superbpe} adds a second BPE stage in which merges may cross spaces, and reported fewer tokens and better English models at a fixed vocabulary. Closest to our goal, MUTANT~\cite{rana2025mutant} builds a tokenizer for 22 Indian languages, English and code with language-aware pre-tokenization and multi-word aware training; its tokenizer was not publicly available for us to measure. Schmidt et al.~\cite{schmidt2024more} showed that fewer tokens do not by themselves give a better model; we therefore make claims about compression only.

\section{Method}
\subsection{Training data}
All training text is human-written web, educational or code text; none was generated by a language model. For about 195 languages we sample FineWeb-2; for 22 languages of India we combine Sangraha~\cite{khan2024indicllmsuite} and FineWeb-2; English comes from FineWeb-Edu. We add the romanized Hindi and Marathi, mathematics and permissively licensed Python sources of our earlier 32,768-token tokenizer (Havon v2), including L3Cube HingCorpus~\cite{nayak2022hingcorpus} and MeCorpus~\cite{chavan2023mecorpus}. Per-language character budgets were 30 million for each of the 22 languages of India with a Sangraha source, 10 million for each of about 56 major world languages, 2 million for every other language, and 120 million for English. The other six FLORES-200 languages of India (Awadhi, Bhojpuri, Chhattisgarhi, Magahi, Mizo and Kashmiri in Devanagari) therefore had the 2 million budget; v3, v4 and v5 still need fewer tokens than GPT-4o on all six. The full sample has 1.69 billion characters; to fit BPE training into 32~GB of memory, the first 60\% of each language's sample (about 1.0 billion characters) was used.

\subsection{Pre-tokenization}
GPT-2's pre-tokenization rule treats Unicode combining marks (category M), which include Devanagari and other Indic vowel signs, as non-letters and splits words at them. Our rule keeps letters and their marks together. Numbers are split into single digits. Scripts written without spaces (Han, kana, Thai, Lao, Khmer, Myanmar and Tibetan) would otherwise form chunks as long as a whole sentence; we cut them into chunks of at most 3 Han or kana characters or 6 letters with their marks, which also keeps training within memory. Text is normalized to NFC.

\subsection{Havon v3 and truncation}
v3 is byte-level BPE trained to 262,144 tokens with the Hugging Face tokenizers library. BPE learns merges in a fixed order, so the first $K$ entries of the vocabulary and the merges that produce them form exactly the tokenizer that training to size $K$ would give; we verified on a small training run that the cut and the directly trained tokenizer have identical vocabularies, merges and encodings, and cut 65,536 to 200,019-token versions from the one run.

\subsection{Havon v4: multi-word merges}
Every tokenizer above, and GPT-4o's, forbids tokens from crossing a space, so ``of the'' costs at least two tokens. Following SuperBPE~\cite{liu2025superbpe}, v4 starts from the first $M$ of v3's 261,824 merges and resumes training with a stage-2 pre-tokenizer that no longer splits at spaces, so later merges can join words. A small share of the inherited merges did not carry over into the final files: 177 of 160,000 (0.11\%) for v4-200k, 343 of 210,000 (0.16\%) for v4-512k and 743 of 160,000 (0.46\%) for v5-200k, mostly rare pieces such as runs of non-breaking spaces, plus some merges of newlines (13 to 34\%, depending on the tokenizer). We have not yet established why the stage-2 implementation drops them. Our stage-2 rule still separates digits and punctuation runs and keeps v3's chunking for scripts without spaces. Stage 2 learned from the first 12\% of each language's v3 sample (335~MB of text), the largest share that fitted in 32~GB of memory, using the open-source implementation released with SuperBPE; no pretrained tokenizer or model was used. We trained two runs:
\begin{itemize}
\item \textbf{v4-200k}: $M=160{,}000$, trained to 200,019 tokens, the size of GPT-4o's o200k\_base.
\item \textbf{v4-512k}: $M=210{,}000$, trained to 524,288 tokens. Its first 262,144 tokens form \textbf{v4-256k}, the size of v3 and Sarvam-30B/105B.
\end{itemize}

\subsection{Evaluation}
We count tokens on FLORES-200~\cite{nllb2022,goyal2022flores} for all 204 languages of the original release, encoding each sentence alone without special tokens. The vocabulary size of v3 and the choice between v3 and v3.1 (Section~\ref{sec:budget}) were made on the dev split (997 sentences); the merge splits of v4 were fixed in advance. Every other number below is on devtest (1,012 sentences), which was not used for any choice. For each language we report the candidate's token count divided by a rival's on the same sentences; below 1 means fewer tokens. Rivals are o200k\_base (GPT-4o), cl100k\_base (GPT-4), and the Hugging Face tokenizers of Sarvam-30B/105B, Sarvam-1, Qwen2.5 and DeepSeek-V3 at pinned revisions.

\section{Results}
\subsection{Vocabulary size for v3}
On dev, compression improved steadily with size (Table~\ref{tab:sweep}), so selection only confirmed the largest size we could train. At 65,536 tokens v3 already needs fewer tokens than GPT-4o on 50 languages; at 262,144 it does on 170 of 204 on dev and 172 on devtest.

\begin{table}[t]
\centering
\caption{Havon v3 at prefix sizes, FLORES-200 dev, relative to GPT-4o.}
\label{tab:sweep}
\footnotesize
\begin{tabular}{rccc}
\toprule
Vocabulary & English & Median (204) & Fewer tokens (of 204) \\
\midrule
65,536 & 1.24 & 1.11 & 50 \\
131,072 & 1.14 & 1.01 & 97 \\
196,608 & 1.10 & 0.96 & 130 \\
262,144 & 1.07 & 0.92 & 170 \\
\bottomrule
\end{tabular}
\end{table}

\subsection{Main results}
Table~\ref{tab:langs} gives the per-language results. Across the 25 languages of India, every Havon tokenizer needs fewer tokens than GPT-4o on every language, with medians from 0.74 (v3 and v4-200k) to 0.65 (v4-512k). Against Sarvam-30B/105B, the strongest public tokenizer for these languages, v3 needs fewer tokens on 22 of 25 (median 0.91), v4-200k and v4-256k on 23 (medians 0.89 and 0.86), and v4-512k on all 25. Bengali and Santali are the two languages where v4-200k and v4-256k still trail Sarvam-30B/105B.

\begin{table*}[t]
\centering
\caption{Token count of each Havon tokenizer divided by the rival's on the same FLORES-200 devtest sentences (below 1: fewer tokens than the rival). Bold: more tokens than the rival. Vocabulary sizes: v3 262,144; v4-200k 200,019 (equal to GPT-4o's); v4-256k 262,144 (equal to Sarvam-30B/105B's); v4-512k 524,288.}
\label{tab:langs}
\footnotesize
\setlength{\tabcolsep}{3.5pt}
\begin{tabular}{lcccc|cccc}
\toprule
 & \multicolumn{4}{c|}{vs GPT-4o (o200k\_base)} & \multicolumn{4}{c}{vs Sarvam-30B/105B} \\
Language & v3 & v4-200k & v4-256k & v4-512k & v3 & v4-200k & v4-256k & v4-512k \\
\midrule
Hindi & 0.85 & 0.77 & 0.73 & 0.66 & \textbf{1.01} & 0.91 & 0.87 & 0.78 \\
Nepali & 0.74 & 0.75 & 0.72 & 0.65 & 0.78 & 0.78 & 0.75 & 0.68 \\
Urdu & 0.86 & 0.79 & 0.76 & 0.68 & 0.93 & 0.86 & 0.82 & 0.74 \\
Awadhi & 0.79 & 0.78 & 0.75 & 0.69 & 0.91 & 0.89 & 0.86 & 0.80 \\
Sindhi (Arabic) & 0.78 & 0.73 & 0.70 & 0.63 & 0.65 & 0.61 & 0.59 & 0.53 \\
Magahi & 0.79 & 0.77 & 0.75 & 0.68 & 0.91 & 0.89 & 0.86 & 0.79 \\
Bengali & 0.74 & 0.76 & 0.72 & 0.65 & \textbf{1.04} & \textbf{1.06} & \textbf{1.01} & 0.90 \\
Chhattisgarhi & 0.80 & 0.79 & 0.76 & 0.71 & 0.92 & 0.91 & 0.88 & 0.82 \\
Maithili & 0.74 & 0.72 & 0.69 & 0.63 & 0.79 & 0.76 & 0.74 & 0.67 \\
Bhojpuri & 0.79 & 0.78 & 0.74 & 0.68 & 0.88 & 0.86 & 0.83 & 0.76 \\
Gujarati & 0.79 & 0.79 & 0.74 & 0.66 & 0.93 & 0.93 & 0.88 & 0.78 \\
Mizo & 0.89 & 0.84 & 0.82 & 0.75 & 0.89 & 0.84 & 0.82 & 0.75 \\
Marathi & 0.68 & 0.69 & 0.66 & 0.60 & 0.88 & 0.88 & 0.85 & 0.77 \\
Telugu & 0.71 & 0.73 & 0.69 & 0.61 & 0.93 & 0.96 & 0.91 & 0.80 \\
Malayalam & 0.76 & 0.80 & 0.76 & 0.67 & 0.91 & 0.95 & 0.90 & 0.79 \\
Tamil & 0.71 & 0.73 & 0.70 & 0.61 & 0.95 & 0.98 & 0.93 & 0.82 \\
Kannada & 0.72 & 0.73 & 0.70 & 0.62 & 0.93 & 0.95 & 0.90 & 0.80 \\
Assamese & 0.66 & 0.67 & 0.63 & 0.57 & 0.65 & 0.66 & 0.63 & 0.56 \\
Sanskrit & 0.66 & 0.66 & 0.64 & 0.59 & 0.77 & 0.77 & 0.74 & 0.68 \\
Kashmiri (Deva.) & 0.83 & 0.85 & 0.81 & 0.76 & 0.93 & 0.96 & 0.92 & 0.86 \\
Kashmiri (Arabic) & 0.72 & 0.74 & 0.71 & 0.65 & 0.79 & 0.81 & 0.77 & 0.71 \\
Punjabi & 0.57 & 0.53 & 0.51 & 0.45 & 0.95 & 0.89 & 0.85 & 0.75 \\
Meitei (Bengali) & 0.47 & 0.47 & 0.45 & 0.41 & 0.57 & 0.58 & 0.55 & 0.50 \\
Odia & 0.27 & 0.28 & 0.27 & 0.24 & 0.94 & 0.98 & 0.94 & 0.83 \\
Santali (Ol Chiki) & 0.14 & 0.15 & 0.14 & 0.12 & \textbf{1.02} & \textbf{1.10} & \textbf{1.03} & 0.93 \\
\midrule
English & \textbf{1.07} & 0.94 & 0.90 & 0.83 & \textbf{1.06} & 0.93 & 0.90 & 0.82 \\
German & \textbf{1.19} & \textbf{1.21} & \textbf{1.16} & \textbf{1.04} & \textbf{1.17} & \textbf{1.19} & \textbf{1.13} & \textbf{1.02} \\
French & \textbf{1.15} & \textbf{1.10} & \textbf{1.05} & 0.95 & \textbf{1.11} & \textbf{1.06} & \textbf{1.01} & 0.91 \\
Spanish & \textbf{1.14} & \textbf{1.07} & \textbf{1.03} & 0.93 & \textbf{1.15} & \textbf{1.08} & \textbf{1.03} & 0.94 \\
Portuguese & \textbf{1.15} & \textbf{1.11} & \textbf{1.07} & 0.97 & \textbf{1.11} & \textbf{1.08} & \textbf{1.03} & 0.93 \\
Russian & \textbf{1.12} & \textbf{1.15} & \textbf{1.09} & 0.97 & \textbf{1.15} & \textbf{1.19} & \textbf{1.13} & \textbf{1.005} \\
Chinese (Simpl.) & \textbf{1.06} & \textbf{1.11} & \textbf{1.07} & 0.99 & \textbf{1.22} & \textbf{1.27} & \textbf{1.23} & \textbf{1.13} \\
Japanese & 0.93 & 0.98 & 0.95 & 0.88 & 0.35 & 0.37 & 0.35 & 0.33 \\
Arabic & \textbf{1.03} & \textbf{1.07} & \textbf{1.02} & 0.91 & 0.96 & 0.99 & 0.95 & 0.85 \\
\midrule
Median, 25 languages of India & 0.74 & 0.74 & 0.71 & 0.65 & 0.91 & 0.89 & 0.86 & 0.78 \\
Median, all 204 languages & 0.92 & 0.92 & 0.89 & 0.81 & 0.88 & 0.87 & 0.84 & 0.77 \\
Fewer tokens, of 25 (India) & 25 & 25 & 25 & 25 & 22 & 23 & 23 & 25 \\
Fewer tokens, of 204 & 172 & 161 & 182 & 203 & 186 & 187 & 192 & 199 \\
\bottomrule
\end{tabular}
\end{table*}

On English, v3 needs 1.07 times GPT-4o's tokens. Adding multi-word merges reverses this: 0.94 for v4-200k, 0.90 for v4-256k and 0.83 for v4-512k. Across all 204 languages, v4-256k needs fewer tokens than GPT-4o on 182 and v4-512k on 203; the one exception for v4-512k is German (1.04). Against cl100k\_base, Sarvam-1, Qwen2.5 and DeepSeek-V3, v4-256k needs fewer tokens on 199 to 204 languages.

\subsection{Equal vocabulary size}
A larger vocabulary needs fewer tokens by itself, and v4-256k's vocabulary is 31\% larger than GPT-4o's. Table~\ref{tab:equal} therefore compares at exactly 200,019 tokens. Cutting v3 to this size raises its English ratio to 1.09 and lowers its wins over GPT-4o to 133. v4-200k, with the same vocabulary size, needs 0.94 of GPT-4o's tokens on English and fewer tokens on 161 of 204 languages. A sentence-level bootstrap (1,000 resamples, 95\% intervals) makes 160 of these 161 wins significant, including all 25 languages of India; the English ratio is 0.939 (0.933 to 0.946). Against v3 at the same size, v4-200k needs 14\% fewer tokens on English, 12\% fewer on Hindi and 3\% fewer on Marathi, with a median reduction of 2.4\% across the 204 languages. This isolates the effect of the multi-word stage from vocabulary size, although the two runs differ in data for the last merges (Section~\ref{sec:lim}).

\begin{table}[t]
\centering
\caption{All three tokenizers have 200,019 tokens. FLORES-200 devtest, relative to GPT-4o (o200k\_base).}
\label{tab:equal}
\footnotesize
\setlength{\tabcolsep}{3pt}
\begin{tabular}{lcccc}
\toprule
 & & Median, & Median, & Fewer tokens \\
Tokenizer & English & India (25) & all 204 & (of 204) \\
\midrule
GPT-4o (o200k\_base) & 1.00 & 1.00 & 1.00 & n/a \\
Havon v3, cut to 200,019 & 1.09 & 0.77 & 0.95 & 133 \\
Havon v4-200k & 0.94 & 0.74 & 0.92 & 161 \\
\bottomrule
\end{tabular}
\end{table}

\subsection{Against a common English baseline}
Ratios against each tokenizer's own English count flatter tokenizers that are weak on English. Table~\ref{tab:common} divides every count by GPT-4o's English count. Sarvam-30B/105B and our earlier v2 need 1.31 to 1.41 times GPT-4o's English tokens to write Marathi and Hindi. v4-256k needs 1.20 and 1.15, and v4-512k 1.09 and 1.03, so at that size Hindi costs almost what English costs on GPT-4o.

\begin{table}[t]
\centering
\caption{FLORES-200 devtest token counts divided by GPT-4o's English count on the same 1,012 sentences. Lower is better.}
\label{tab:common}
\footnotesize
\begin{tabular}{lrccc}
\toprule
Tokenizer & Vocab & English & Marathi & Hindi \\
\midrule
GPT-4o (o200k) & 200,019 & 1.00 & 1.82 & 1.57 \\
Sarvam-30B/105B & 262,144 & 1.01 & 1.41 & 1.32 \\
Sarvam-1 & 68,096 & 1.17 & 1.26 & 1.32 \\
Havon v2 & 32,768 & 1.19 & 1.31 & 1.41 \\
Havon v3 & 262,144 & 1.07 & 1.24 & 1.33 \\
Havon v4-200k & 200,019 & 0.94 & 1.25 & 1.20 \\
Havon v4-256k & 262,144 & 0.90 & 1.20 & 1.15 \\
Havon v4-512k & 524,288 & 0.83 & 1.09 & 1.03 \\
\bottomrule
\end{tabular}
\end{table}

\subsection{Controlled test of the multi-word stage}
To measure the multi-word stage on identical text, we trained a 40,000-token tokenizer on the same sample twice: once with ordinary BPE, and once inheriting the first 32,000 merges and finishing with stage 2. Stage 2 used a pre-tokenizer without chunking for scripts without spaces, which none of the six test languages uses. On devtest, stage 2 cut token counts by 8.8\% for English, 14.2\% for Hindi, 10.8\% for French, 10.5\% for Spanish, 5.0\% for German and 3.0\% for Marathi. Hindi probably gains most because many short function words (postpositions and auxiliaries such as \textit{ka}, \textit{ki}, \textit{hai}) are written as separate words; Marathi attaches many of these to the noun or verb, so its words are already long.

\subsection{Data budgets are not a substitute}
\label{sec:budget}
Before adding multi-word merges we tried rebalancing v3's data toward English and the major world languages (v3.1: 200 million English characters, 15 million for each major language, 20 million for each language of India and 1.5 million for the rest). On dev, it moved the English ratio only from 1.07 to 1.06 and lowered the number of languages where the tokenizer needed fewer tokens than GPT-4o from 170 to 162, so we kept v3's budgets.

\subsection{Where v4 does not help}
At 262,144 tokens, v4 needs more tokens than v3 on 26 of 204 languages, by at most 3.5\%, among them Dzongkha, Burmese, Tibetan, Khmer, Lao, Thai and Japanese. v4 gives up v3's last 51,824 subword merges to make room for multi-word merges, and these scripts, which are written without spaces and chunked, gain little from merges across spaces. Against v3 at the same 200,019 tokens, v4-200k is worse on 26 languages, most of all Khmer (15\%), Burmese, Dzongkha and Tibetan. Against GPT-4o, v4-200k still needs more tokens on 43 languages, mainly German (1.21), Russian (1.15), Dutch, Tajik, Portuguese, Belarusian, Chinese (1.11), French (1.10) and Spanish (1.07), whose budgets were small next to English.

\subsection{Havon v5: code}
\label{sec:v5}
v4's code sample was 40 million characters of Python only. Measured with fixed seeds on up to 1.5~MB of text per category (0.27~MB for Rust and 0.36~MB for shell scripts, the rarest in the sample), it needs 1.11 times GPT-4o's tokens on Python and 1.19 to 1.29 times on nine other programming languages (Table~\ref{tab:v5}). Havon v5 keeps v4's recipe and settings but replaces that sample with 120.1 million characters of permissively licensed code in 15 languages (from The Stack~v3, files whose own license is permissive; about 8 million characters per language), retrains the subword stage on the first 60\% of each sample, and runs the same multi-word stage ($M=160{,}000$, 200,019 tokens, 12\% of each sample). On code, v5 needs 1.03 to 1.22 times GPT-4o's tokens; the largest gains are Java (1.28 to 1.09), JavaScript (1.21 to 1.08), TypeScript (1.20 to 1.04) and Go (1.19 to 1.03), and SQL (1.22) and shell scripts (1.14) remain the weakest. Mathematics and StackExchange text barely change. On FLORES-200, v5 needs fewer tokens than GPT-4o on 160 of 204 languages, 155 of them significant under the same bootstrap, and on all 25 languages of India (all significant, median 0.742 against v4's 0.739); its English ratio is 0.941 (0.934 to 0.947).

\begin{table}[t]
\centering
\caption{Tokens relative to GPT-4o (o200k\_base) at 200,019 tokens. Code, mathematics and StackExchange: up to 1.5~MB of text each (Rust 0.27~MB, shell 0.36~MB, C 0.55~MB, SQL 0.69~MB). The code comes from Stack~v3 files not used for training and the StackExchange text from a dataset not used for training; the mathematics text comes from the first file of each source, where the mathematics training sample also starts, so it may partly overlap it. FLORES-200 devtest for languages.}
\label{tab:v5}
\footnotesize
\setlength{\tabcolsep}{4pt}
\begin{tabular}{lcc}
\toprule
Text & Havon v4-200k & Havon v5-200k \\
\midrule
Python & 1.11 & 1.10 \\
JavaScript & 1.21 & 1.08 \\
TypeScript & 1.20 & 1.04 \\
Java & 1.28 & 1.09 \\
C / C++ & 1.23 / 1.20 & 1.09 / 1.08 \\
Go & 1.19 & 1.03 \\
Rust & 1.21 & 1.11 \\
SQL & 1.29 & 1.22 \\
Shell & 1.26 & 1.14 \\
Maths (FineMath / OpenWebMath) & 1.03 / 0.97 & 1.02 / 0.97 \\
StackExchange (3 sites) & 0.90 & 0.90 \\
English (FLORES-200) & 0.94 & 0.94 \\
India, median of 25 (FLORES-200) & 0.739 & 0.742 \\
Fewer tokens than GPT-4o (of 204) & 161 & 160 \\
\bottomrule
\end{tabular}
\end{table}

\subsection{A fixed vocabulary is a budget}
To test how far reallocation alone can go, we trained a variant (v5.1) that re-drew the samples of the 44 languages on which v5 still needed more tokens than GPT-4o, at three times their characters, and doubled SQL, shell and Rust code. At the same 200,019 tokens, it needs fewer tokens than GPT-4o on 185 of 204 languages: it gains 33 languages but loses 8 that v5 had won (among them Greek, Finnish, Korean, Polish and Swedish), its median on the 25 languages of India rises from 0.742 to 0.770, its English ratio from 0.941 to 0.957, and the extra SQL, shell and Rust code changes those languages by less than 0.5\%. At a fixed size, each language's compression is paid for by another's; together with v4-512k's 203 of 204 at 524,288 tokens, this suggests that matching GPT-4o on every language at its own size would cost part of the advantage on languages of India. We release v5, not v5.1.

\subsection{Comparison with other public tokenizers}
\begin{figure*}[t]
\centering
\includegraphics[width=0.82\textwidth]{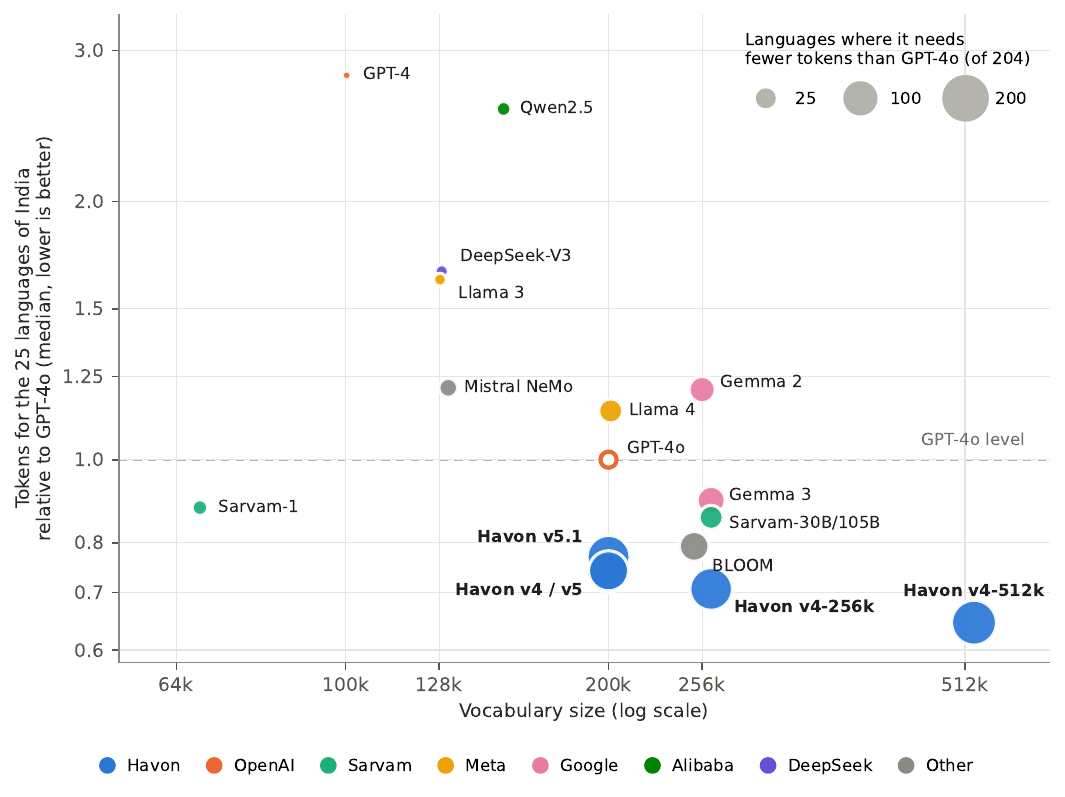}
\caption{Tokens needed for the 25 languages of India relative to GPT-4o (median over the languages; lower is better) against vocabulary size, FLORES-200 devtest. Bubble area: number of the 204 FLORES-200 languages on which the tokenizer needs fewer tokens than GPT-4o. Havon v4-200k and v5-200k overlap (0.739 and 0.742).}
\label{fig:tokenizers}
\end{figure*}

Fig.~\ref{fig:tokenizers} places every tokenizer we measured by vocabulary size and by its median token count on the 25 languages of India relative to GPT-4o. Besides the rivals listed in the evaluation setup, we measured Llama~3 and Llama~4, Gemma~2 and Gemma~3, BLOOM and Mistral NeMo, from public copies of their tokenizer files on Hugging Face (NousResearch/Meta-Llama-3-8B; unsloth/Llama-4-Scout-17B-16E, gemma-2-9b, gemma-3-4b-pt and Mistral-Nemo-Base-2407; bigscience/bloom-560m), downloaded on 8 November 2026; their SHA-256 hashes are in the repository. Their medians range from 0.79 (BLOOM) and 0.90 (Gemma~3) to 1.62 (Llama~3); every Havon tokenizer from 200,019 tokens upwards is lower (0.65 to 0.77), and none of the other tokenizers needs fewer tokens than GPT-4o on more than 84 of the 204 languages.

\subsection{Romanized text and morphology}
On PHINC~\cite{srivastava2020phinc}, 13,732 Hinglish social media posts with human English translations that were not in our training data, v4-256k needs 1.39 tokens per word on Hinglish and 1.26 on English, against 1.61 and 1.36 for GPT-4o and 1.53 and 1.38 for Sarvam-30B/105B~\cite{johny2026tax}. On the MorphTok Hindi and Marathi segmentations~\cite{brahma2025morphtok}, v4-256k's boundary F1 is 0.40 and 0.52, about the same as v3 (0.41 and 0.52), against 0.31 and 0.43 for GPT-4o and 0.48 and 0.48 for Sarvam-30B/105B: multi-word merges do not change word-internal boundaries much. These measurements use v4-256k; v4-200k was not measured on them.

\section{Discussion}
Two levers did most of the work. A pre-tokenizer that respects Indic combining marks and a vocabulary with dedicated budgets for every language of India give large reductions on the languages GPT-4o serves worst: v3 needs 0.14 of GPT-4o's tokens on Santali and 0.27 on Odia. Those budgets cost English and European compression at a fixed size, and shifting budgets only trades one language for another. The multi-word stage recovers English and the Romance languages without taking back the Indic gains, and is the only change that moved v4 below GPT-4o on English at an equal vocabulary size. The remaining gaps are German, Russian and Chinese, where GPT-4o's training data presumably has far more text than our 10 million character budgets.

For a model builder, a 200,019-token tokenizer that needs 6\% fewer tokens than GPT-4o's on English and about a quarter fewer on the median language of India means more text in the same context and fewer decoding steps for the same output. Whether a model learns as well from these tokens is a separate question: SuperBPE's authors report better English models with multi-word tokens~\cite{liu2025superbpe}, but we have not yet evaluated a model trained with v4 or v5.

\section{Limitations}
\label{sec:lim}
(1) All results are token counts; we have not evaluated a model trained with v3, v4 or v5, and fewer tokens do not imply a better model~\cite{schmidt2024more}. (2) The training sample is FineWeb-2 and Sangraha web text. A check for FLORES-200 sentences in it (13-word n-grams, or 30-character n-grams for scripts written without spaces; a sentence counts if half its n-grams occur) found none in the 188 languages that have a sample, apart from two one-word lines; paraphrases and translations are not detected by such a check. (3) The multi-word stage learned from 12\% of the sample to fit in memory, so the last merges of v4 see less text than its subword merges; a machine with more memory could use more. (4) v4-200k inherits 160,000 merges and v4-256k 210,000; we did not tune this split. (5) FLORES-200 is formal, Wikipedia-style text; conversational text was not measured, code and mathematics were measured on 0.27 to 1.5~MB per category, and the romanized and morphological measurements cover only Hindi and Marathi. (6) We compare against public tokenizers only; Claude and Gemini were excluded because their tokenizers are not public. Llama~3 and~4, Gemma~2 and~3 and Mistral NeMo were read from public copies of their tokenizer files rather than the gated official repositories; we did not verify that these copies match the official files byte for byte. (7) Part of the training data (L3Cube HingCorpus and MeCorpus) is licensed CC BY-NC-SA 4.0 for research, so the Havon tokenizer files are released under the same license; HingCorpus was read from a third-party Hugging Face mirror (hinglish-orthobench/HingCorpus).

\section{Conclusion}
A tokenizer trained from scratch on human-written text in about 200 languages, with multi-word merges added in a second stage, needs fewer tokens than GPT-4o's on English and on 161 of 204 FLORES-200 languages at exactly GPT-4o's vocabulary size, and on all 25 languages of India by a median of about a quarter. At 524,288 tokens it needs fewer on 203 of 204. Havon v5 extends the same recipe to code while keeping these results. Next, we will train and evaluate models with v5 to test whether the lower token counts carry over to quality.

\section*{Reproducibility}
The v3, v4 and v5 tokenizer files (with SHA-256 hashes), the training scripts (\texttt{train\_v3.py}, \texttt{superbpe\_stage2.py}) with the per-language budgets and data sources, the truncation (\texttt{truncate.py}) and evaluation (\texttt{eval\_world.py}, \texttt{eval\_domains.py}) scripts, the devtest token counts for every tokenizer in this paper and the dev counts used for selection, the bootstrap intervals (\texttt{bootstrap\_ci.py}) and the FLORES-200 overlap check (\texttt{contamination.py}) with their results, and the data and script behind Fig.~\ref{fig:tokenizers} are public at \url{https://github.com/HavonAI/indic-token-tax}. The v3.1 and v5.1 tokenizers are not released; their token counts are. Stage 2 uses the SuperBPE fork of the Hugging Face tokenizers library at commit 757f2a5.

\end{document}